\documentclass[12pt]{article}
\usepackage{chicago}
\newtheorem{THEOREM}{Theorem}[section]
\newenvironment{theorem}{\begin{THEOREM} \hspace{-.85em} {\bf :} }%
                        {\end{THEOREM}}
\newtheorem{LEMMA}[THEOREM]{Lemma}
\newenvironment{lemma}{\begin{LEMMA} \hspace{-.85em} {\bf :} }%
                      {\end{LEMMA}}
\newtheorem{COROLLARY}[THEOREM]{Corollary}
\newenvironment{corollary}{\begin{COROLLARY} \hspace{-.85em} {\bf :} }%
                          {\end{COROLLARY}}
\newtheorem{PROPOSITION}[THEOREM]{Proposition}
\newenvironment{proposition}{\begin{PROPOSITION} \hspace{-.85em} {\bf :} }%
                            {\end{PROPOSITION}}
\newtheorem{DEFINITION}[THEOREM]{Definition}
\newenvironment{definition}{\begin{DEFINITION} \hspace{-.85em} {\bf :} \rm}%
                            {\end{DEFINITION}}
\newtheorem{CLAIM}[THEOREM]{Claim}
\newenvironment{claim}{\begin{CLAIM} \hspace{-.85em} {\bf :} \rm}%
                            {\end{CLAIM}}
\newtheorem{EXAMPLE}[THEOREM]{Example}
\newenvironment{example}{\begin{EXAMPLE} \hspace{-.85em} {\bf :} \rm}%
                            {\end{EXAMPLE}}
\newtheorem{REMARK}[THEOREM]{Remark}
\newenvironment{remark}{\begin{REMARK} \hspace{-.85em} {\bf :} \rm}%
                            {\end{REMARK}}
\newcommand{\thm}{\begin{theorem}}
\newcommand{\lem}{\begin{lemma}}
\newcommand{\pro}{\begin{proposition}}
\newcommand{\dfn}{\begin{definition}}
\newcommand{\rem}{\begin{remark}}
\newcommand{\xam}{\begin{example}}
\newcommand{\cor}{\begin{corollary}}
\newcommand{\ethm}{\end{theorem}}
\newcommand{\elem}{\end{lemma}}
\newcommand{\epro}{\end{proposition}}
\newcommand{\edfn}{\bbox\end{definition}}
\newcommand{\erem}{\bbox\end{remark}}
\newcommand{\exam}{\bbox\end{example}}
\newcommand{\ecor}{\end{corollary}}
\newcommand{\eprf}{\bbox\vspace{0.1in}}
\newcommand{\beqn}{\begin{equation}}
\newcommand{\eeqn}{\end{equation}}
\newcommand{\bbox}{\vrule height7pt width4pt depth1pt}

\newcommand{\clm}{\begin{claim}}
\newcommand{\eclm}{\end{claim}}
\newcommand{\sat}{\models}

\newcommand{\FF}{{\bf F}}

\renewcommand{\phi}{\varphi}
\newcommand{\F}{{\cal F}}

\newcommand{\K}{{\cal K}}

\newcommand{\R}{{\cal R}}

\newcommand{\U}{{\cal U}}
\newcommand{\V}{{\cal V}}

\newcommand{\ol}{\setlength{\itemsep}{0pt}\begin{enumerate}}
\newcommand{\eol}{\end{enumerate}\setlength{\itemsep}{-\parsep}}
\newcommand{\ul}{\setlength{\itemsep}{0pt}\begin{itemize}}
\newcommand{\dl}{\setlength{\itemsep}{0pt}\begin{description}}
\newcommand{\edl}{\end{description}\setlength{\itemsep}{-\parsep}}
\newcommand{\eul}{\end{itemize}\setlength{\itemsep}{-\parsep}}

\newcommand{\commentout}[1]{}

\newcommand{\bi}{\begin{itemize}}
\newcommand{\ei}{\end{itemize}}
\newcommand{\be}{\begin{enumerate}}
\newcommand{\ee}{\end{enumerate}}

\newcommand{\Scal}{{\cal S}}
\newcommand{\ML}{\mathit{MD}}
\newcommand{\intension}[1]{[\![ #1 ]\!]}
\renewcommand{\FF}{\mathit{FF}}

\begin{document}

\title{Mathematical Explanations}
\author{
  %Jordan Ellenberg\\
%  Mathematics Department\\
%  University of Wisconsin, Madison\\
%  ellenber@math.wisc.edu
%  \and
Joseph Y. Halpern\thanks{Supported in part by NSF grants IIS-178108
  and IIS-1703846, MURI grant W911NF-19-1-0217, and ARO grant W911NF-22-1-0061.}\\
Computer Science Department\\
Cornell University\\
halpern@cs.cornell.edu}
\date{ }
\maketitle

\begin{abstract}
A definition of what counts as an explanation of mathematical
statement, and when one explanation is better than another, is given.
Since all mathematical facts must be true in all causal models, and
hence known by an agent, mathematical facts cannot
be part of an explanation (under the standard notion of explanation).
This problem is solved using impossible possible worlds.
\end{abstract}

\section{Introduction}

Consider two mathematical questions like ``Why is 4373 the sum of two
squares?'' or ``For $f(x) = x^{11} - 6x^{10} + 11x^9 -17x^8 +22x^7 -
5x^6 + 10x^5 + x^9 - 2x^3 -x +2$, why is $f(2) = 0$?''.
An explanation for the first question might simply be a demonstration:
$4373 = 3844 + 529 = 62^2 + 23^2$.  An arguably better explanation, at least for
those who know Fermat's two-squares theorem, which says that an odd
prime is congruent to 1 mod 
4 (i.e., has a remainder of 1 when divided by 4) if and only if it is
the the sum of two
squares, is the observation that 4373 is 1 mod 4 (since $4373 = 4
\times 1093 + 1$) and is prime.  Similarly, an explanation for the
second question is just the observation that (after some laborious
multiplication and addition) that
\begin{equation}\label{eq:factor}
  2^{11} - 6 \times 2^{10} + 11 \times 2^9 -17 \time 2^8 +22 \times 2^7 -
  5 \times 26 + 10 \times 2^5 + 2^9 - 2 \times 2^3 -2 +2 = 0;
\end{equation}
  a better
explanation (at least for many) is the observation that
\begin{equation}\label{eq1}
  f(x) = (x-2)(x^{10} - 4 x^9 + 3 x^8 = 11x^7 - 5x^5 + x^3 -1).
 \end{equation}

What makes these explanations?  Why is one explanation better than
another?  There have been many definitions of explanation proposed in
the literature.  Hempel's \citeyear{Hempel65} {\em
  deductive-nomological\/} model does a good job of accounting for why
these count as explanations.  In this model, an explanation consists
of a ``law of nature'' and some additional facts that together
imply the \emph{explanandum} (the fact to be explained).  In the first
example, the law of nature is Fermat's two-squares theorem;
the additional facts are the
observation that 173 is a prime and that it is congruent to 1 mod 4.
Similarly, in the second example, the law of nature is the fact that
if $(x-m)$ is a factor of a polynomial $f$ whose coefficients are
integers (i.e., $f(x) = (x-2)g(x)$, where $g$ is a polynomial whose
coefficients are integers), then $f(2) = 0$, together with
(\ref{eq1}).

As is well known, there are difficulties with Hempel's account of
explanation.  For one thing, it does not take causality into account;
for another, it does not take into account the well-known observation
that what counts as an explanation is relative to what an
agent knows \cite{Gardenfors1,Salmon84}.
As G\"ardenfors \citeyear{Gardenfors1} observes, an agent seeking 
an explanation of why Mr.\ Johansson has been taken ill with lung cancer
will not consider the fact that he worked for years in asbestos
manufacturing an explanation if he already knew this fact.
%In our example, the theorem
%that primes congruent to 1 mod 4 can be written as the sum of two
%squares is not part of the explanation if the agent already knows that.
Finally, this definition does not give us a way to say that one
explanation is better than another.

In this paper, we focus on the definition of (causal)
explanation given by Halpern and Pearl \cite{Hal48,HP01a}.
It has been shown to do quite well on many problematic examples for
which other definitions of explanations have difficulty.  The
definition starts with a definition of causality in a causal model.  
In a causal model, we can talk about one event being a cause of
another. Explanation is then defined relative to an agent's
\emph{epistemic state}, which, roughly speaking, consists of a set of
casual models and a probability on them.  The epistemic state can be
thought of as describing the agent's beliefs about how causality works
in the world; an agent is said to know a fact if the fact has probability 1
according to his epistemic state. Following G\"ardenfors, we would
expect an explanation to be something that the agent did
not already know.

But now we see a problem: all mathematical facts must be true in all
causal models, and hence must be known by an agent.  That means
mathematical facts cannot be part of an explanation.  On the other
hand, how can we give an explanation of a mathematical statement
without invoking facts of mathematics?

In this note, we sketch a solution to this problem using 
the Halpern-Pearl (HP) definition of explanation; we expect that the
idea might well apply to other definitions as well.  The solution
takes as its point of departure the notion of ``impossible'' possible
worlds.  This idea has a long history in epistemic logic
\cite{Cress,Cress2,Cress3,Hi2,Kr2,Rant}.  To understand it, recall
that Hintikka \citeyear{Hi1} assumed that an agent considered a number
of worlds possible, and took the statement 
``Agent $a$ knows $\phi$'' to be true if $\phi$ was true in all the
worlds that an agent considered possible.  
So one way to model the fact that $a$ knows it's sunny in Ithaca and 
doesn't know whether it's sunny in Berkeley is by saying $a$ considers
two worlds possible; in one, it's sunny in both Ithaca and Berkeley,
while in the other, it's sunny in Ithaca and raining in Berkeley.

With this viewpoint, ``possibility'' is the dual of knowledge.  Agent $a$
considers $\phi$ possible if $a$ does not know not $\phi$, which is
the case if $a$ considers at least one world possible where $\phi$ is
true.  This approach also runs into trouble with mathematical
statements.  Suppose that agent $a$ is presented with
a 200-digit number $n$.  It seems reasonable to say $a$ doesn't know
whether $n$ is prime; $a$ considers it possible both that $n$is prime and
that $n$ isn't prime.  But suppose that $n$ is in fact prime.  That would
means that the world that $a$ considers possible where $n$ is not prime is
inconsistent with basic number theory.  The ``impossible'' possible
worlds approach referred to above allows agent $a$ to consider such
worlds possible.

Here we show that the analogous approach, when applied to causal
models, allows us to deal with mathematical explanations.  
Specifically, in the case of the sum of squares, the epistemic state
would include causal models where 4373 is not a prime (and/or causal
models where it is not congruent to 1 mod 4); in the case of the
polynomial, the epistemic state would include causal models where
$x-2$ is not a factor of $f$.  This approach will also allow us to say
that one mathematical explanation is better than another.

The rest of this paper is organized as follows: in the next section,
we briefly review causal models and the HP definition of explanation.
In Section~\ref{sec:mathexplanation}, we show how adding
``impossible'' causal models allows us to capture mathematical explanations.

\section{Causal models and the HP definition of explanation}\label{sec:review}

In this section, we review causal models and the HP definition of
explanation.  The reader is encouraged to consult \cite{Hal48}, for
where this material is largely taken (almost verbatim), for further details.
However, it should be possible to understand how we deal with
mathematical explanations without understanding all the details of
these definitions.

\subsection{Causal models}

We assume that the world is described in terms of j
variables and their values.  Some variables may have a causal
influence on others. This influence is modeled by a set of {\em
  structural equations}. It is conceptually useful to split the
variables into two sets: the {\em exogenous\/} variables, whose values
are determined by factors outside the model, and the
{\em endogenous\/} variables, whose values are ultimately determined by
the exogenous variables.  The structural equations describe how these
values are determined.

Formally, a \emph{causal model} $M$
is a pair $(\Scal,\F)$, where $\Scal$ is a \emph{signature}, which explicitly
lists the endogenous and exogenous variables  and characterizes
their possible values, and $\F$ defines a set of \emph{(modifiable)
structural equations}, relating the values of the variables.  
A signature $\Scal$ is a tuple $(\U,\V,\R)$, where $\U$ is a set of
exogenous variables, $\V$ is a set 
of endogenous variables, and $\R$ associates with every variable $Y \in 
\U \cup \V$ a nonempty set $\R(Y)$ of possible values for 
$Y$ (i.e., the set of values over which $Y$ {\em ranges}).  
%For simplicity, we assume here that $\V$ is finite, as is $\R(Y)$ for
%every endogenous variable $Y \in \V$.
$\F$ associates with each endogenous variable $X \in \V$ a
function denoted $F_X$
(i.e., $F_X = \F(X)$)
such that $F_X: (\times_{U \in \U} \R(U))
\times (\times_{Y \in \V - \{X\}} \R(Y)) \rightarrow \R(X)$.
This mathematical notation just makes precise the fact that 
$F_X$ determines the value of $X$,
given the values of all the other variables in $\U \cup \V$.
We typically simplify notation and write $X = Y + U$ instead of 
$F_X(Y,Y',U) = Y+U$.  (The fact that $Y'$ does not appear on the
right-hand side of the equation means that the value of $X$ does not
depend on $Y'$.)

The structural equations define what happens in the presence of external
interventions. Setting the value of some set of variables $\vec{X}$ to $\vec{x}$ in a causal
model $M = (\Scal,\F)$ results in a new causal model, denoted
$M_{\vec{X}\gets \vec{x}}$, which is identical to $M$, except that the
equations for variables in $\vec{X}$ in $\F$ are replaced by $X = x$ for each $X \in \vec{X}$ and its corresponding
value $x \in \vec{x}$.

%The dependencies between variables in a causal model $M = ((\U,\V,\R),\F)$
%can be described using a {\em causal network}\index{causal
%  network} (or \emph{causal graph}),
%whose nodes are labeled by the endogenous and exogenous variables in
%$M$, with one node for each variable in $\U \cup
%\V$.  The roots of the graph are (labeled by)
%the exogenous variables.  There is a directed edge from  variable $X$
%to $Y$ if $Y$ \emph{depends on} $X$; this is the case
A variable $Y$ \emph{depends on} $X$ 
if there is some setting of all the variables in 
$\U \cup \V$ other than $X$ and $Y$ such that varying the value of
$X$ in that setting results in a variation in the value of $Y$; that
is, there is 
a setting $\vec{z}$ of the variables other than $X$ and $Y$ and values
$x$ and $x'$ of $X$ such that
$F_Y(x,\vec{z}) \ne F_Y(x',\vec{z})$.
A causal model  $M$ is \emph{recursive} (or \emph{acyclic})
%if its causal graph is acyclic.
if there is no cycle of dependencies.
It should be clear that if $M$ is an acyclic  causal model,
then given a \emph{context}, that is, a setting $\vec{u}$ for the
exogenous variables in $\U$, the values of all the other variables are
determined (i.e., there is a unique solution to all the equations).
We can determine these values by starting at the top of the graph and
working our way down.
In this paper, following the literature, we restrict to recursive models.
A pair $(M,\vec{u})$ consisting of a causal model $M$ and a
context $\vec{u}$ is called a \emph{(causal) setting}.

A {\em causal formula (over $\Scal$)\/} is one of the form
$[Y_1 \gets y_1, \ldots, Y_k \gets y_k] \phi$,
where
\begin{itemize}
\item
$\phi$ is a Boolean
combination of primitive events,
\item $Y_1, \ldots, Y_k$ are distinct variables in $\V$, and
\item $y_i \in \R(Y_i)$.
\end{itemize}
Such a formula is
abbreviated
as $[\vec{Y} \gets \vec{y}]\phi$.
The special
case where $k=0$
is abbreviated as
$\phi$.
Intuitively,
$[Y_1 \gets y_1, \ldots, Y_k \gets y_k] \phi$ says that
%$\phi(\vec{u})$ holds in the counterfactual world that would arise if
$\phi$ would hold if
$Y_i$ were set to $y_i$, for $i = 1,\ldots,k$.

A causal formula $\psi$ is true or false in a setting.
As usual, we write $(M,\vec{u}) \models \psi$  if
the causal formula $\psi$ is true in
the setting $(M,\vec{u})$.
The $\models$ relation is defined inductively.
$(M,\vec{u}) \models X = x$ if
the variable $X$ has value $x$
in the unique (since we are dealing with acyclic models) solution
to the equations in
$M$ in context $\vec{u}$ (that is, the
unique vector of values for the exogenous variables that simultaneously satisfies all
equations 
in $M$ 
with the variables in $\U$ set to $\vec{u}$).
Finally, 
$(M,\vec{u}) \models [\vec{Y} \gets \vec{y}]\varphi$ if 
$(M_{\vec{Y} = \vec{y}},\vec{u}) \models \varphi$.

\subsection{Actual causality}
A standard use of causal models is to define \emph{actual
causation}: that is, 
what it means for some particular event that occurred to cause 
 another particular event. 
There have been a number of definitions of actual causation given
for acyclic models
(e.g., 
\cite{beckers21c,GW07,Hall07,HP01b,Hal48,hitchcock:99,Hitchcock07,Weslake11,Woodward03}).
%joe2
%Although most of what we say in Section~\ref{sec:exam} applies without
Although most of what we say in the remainder of the paper applies without
change to other definitions of 
actual causality in causal models, for definiteness, we focus here on
%joe10
%what \cite{Hal48} calls the \emph{modified} Halpern-Pearl definition,
what has been called the \emph{modified} Halpern-Pearl definition
\cite{Hal47,Hal48}, 
which we briefly review.
(See \cite{Hal48} for more intuition and motivation.)

The events that can be causes are arbitrary conjunctions of primitive
events (formulas of the form $X=x$); the events that can be caused are arbitrary
Boolean combinations of primitive events.  
The definition takes as its point of departure the notion of
\emph{but-for} causality, widely used in the law; the intuition for
but-for causality is that $A$ is a cause of $B$ if, had $A$ not
occurred, $B$ would not have occurred.
However, as is well known, the but-for test is not always sufficient
to determine causality.  Consider the following well-known example,
taken from \cite{HallP03}:
\begin{quote}
Suzy and Billy both pick up rocks
and throw them at  a bottle.
Suzy's rock gets there first, shattering the
bottle.  Since both throws are perfectly accurate, Billy's would have
shattered the bottle
had it not been preempted by Suzy's throw.
\end{quote}
Here the but-for test fails.  Even if Suzy hadn't thrown, the bottle
would have shattered.  Nevertheless, we want to call Suzy's throw a
cause of the bottle shattering.
The following definition allows for causality beyond but-for causality.

\dfn\label{def:AC}
$\vec{X} = \vec{x}$ rather than $\vec{X} = \vec{x}'$ is 
an 
\emph{actual cause} of $\phi$ rather than $\phi'$ in 
$(M,\vec{u})$ if the
following three conditions hold: 
\begin{description}
\item[{\rm AC1.}]\label{ac1} $(M,\vec{u}) \sat (\vec{X} = \vec{x})$ and 
$(M,\vec{u}) \sat \phi$.
%(That is, for $\vec{X} = \vec{x}$ to be a cause of $\phi$, both $\vec{X}
\item[{\rm AC2.}]\label{ac2} There is a set $\vec{W}$ of variables in $\V$
and a setting $\vec{x}'$ of the variables in $\vec{X}$ such that 
if  $(M,\vec{u}) \sat \vec{W} = \vec{w}$, then
$$(M,\vec{u}) \sat [\vec{X} \gets \vec{x}',
\vec{W} \gets \vec{w}]\neg \phi.$$
\item[{\rm AC3.}] \label{ac3}
$\vec{X}$ is minimal; no subset of $\vec{X}$ satisfies
conditions AC1 and AC2.
\end{description}
\edfn
\vspace{-0.2cm}
\noindent AC1 just says that $\vec{X}=\vec{x}$ cannot
be considered a cause of $\phi$ unless both $\vec{X} = \vec{x}$ and
$\phi$ actually happen.  AC3 is a minimality condition, which says
that a cause has no irrelevant conjuncts.  
AC2 captures the standard
but-for condition ($\vec{X}=\vec{x}$ is a cause of $\phi$ if, had 
$\vec{X}$ beem $\vec{x}'$ rather than $\vec{x}$, $\phi$
would not have happened) but allows us to apply it while keeping fixed
some variables (i.e., the variables in $\vec{W}$) to the value that
they had in the actual  setting $(M,\vec{u})$.
Note that if $\vec{W} = \emptyset$, then we get but-for causality.
Thus, this definition generalizes but-for causality (although, for the
examples in this
paper, but-for causality suffices).

\begin{example}
Suppose that we want to model the fact that
if an arsonist drops a match or lightning strikes then a forest fire
starts.  We could use endogenous
binary variables $\ML$ (which is 1 if the arsonist drops a match, and
0 if he doesn't), $L$ (which is 1 if lightning strikes, and 0 if it
doesn't), and $\FF$ (which is 1 if there is a forest fire, and 0
otherwse), with the equation 
$\FF = \max(L,\ML)$; that is, the value of the variable $\FF$ is the
maximum of the values of the variables $\ML$ and $L$.  This 
equation says, among other things, that if $\ML=0$ and $L=1$, then $\FF=1$.
Alternatively, if we want to model the fact that a fire requires both a
lightning strike \emph{and} a dropped match (perhaps the wood is so wet
that it needs two sources of fire to get going), then the only change in the
model is that the equation for $\FF$ becomes $\FF = \min(L,\ML)$; the
value of $\FF$ is the minimum of the values of $\ML$ and $L$.  The only
way that $\FF = 1$ is if both $L=1$ and $\ML=1$.  

There is also an exogenous variable $U$ that determines whether
the lightning strikes and whether the match is dropped.  
$U$ can take on 
four possible values of the form $(i,j)$, where $i$ and
$j$ are each either 0 or 1.  Intuitively,
$i$ describes whether the external conditions are such that the
lightning strikes (and encapsulates all such conditions, e.g.,
humidity and temperature), and $j$ describes whether 
the arsonist drops the match (and thus encapsulates all the psychological
conditions that determine this).

Consider the context where $U=(1,1)$, so the arsonist drop a match and
the lightning strikes.  
In the \emph{conjunctive} model, where a fire requires both a
lightning strike and a dropped match, it is easy to check that both
$L=$ and  $\ML=1$ are (separately) causes of $\FF=1$.  Indeed, both
$L=1$ and $\ML=1$ are but-for causes.  On the other hand, in the
disjunctive model, where either $L=1$ or $\ML=1$ suffices for the
fire, neither $L=1$ nor $\ML=1$ is a cause (since changing either one
does not result in there not being a fire, no matter what we fix); the
cause of the fire is the conjunction $L=1 \land \ML=1$.  As we shall
see, things go the other way when it comes to explanation.
\end{example}

\subsection{The HP definition of explanation}

Explanation is defined relative to an epistemic state ($\K,\Pr)$, where
$\K$ is a set of causal settings and $\Pr$ is a probability on them.
For simplicity, we assume 
that $\Pr(M,\vec{u}) > 0$ for each causal setting $(M,\vec{u})
\in \K$.  
The epistemic state reflects the agent's uncertainty regarding what
the true causal model is.  The basic definition of explanation does
not make use of $\Pr$, just $\K$.%
\footnote{The definition given here is taken from
\cite{Hal48}, and differs slightly from the original definition given
in \cite{HP01a}.}

We now give the formal definition, and then give intuition for the
clauses, particularly EX1(a)).
\dfn\label{def:explanation1} 
{\em $\vec{X} = \vec{x}$ is an explanation
of $\phi$ relative to a set $\K$ of causal settings\/} 
if the following conditions hold:
\begin{description}
\item[{\rm EX1(a).}] 
If $(M,\vec{u}) \in \K$ and $(M,\vec{u}) \sat \vec{X} = \vec{x} \land
\phi$, then 
  there exists a conjunct $X=x$ of \mbox{$\vec{X} = \vec{x}$} and a (possibly empty)
  conjunction $\vec{Y} = 
  \vec{y}$ such that \mbox{$X = x \land \vec{Y} = \vec{y}$} is
  a cause of $\phi$ in $(M,\vec{u})$.  
\item[{\rm EX1(b).}]  $(M',\vec{u}') \sat [\vec{X} \gets \vec{x}]\phi$ for all
  settings $(M',\vec{u}') \in \K$.  
\item[{\rm EX2.}] $\vec{X}$ is minimal; there is no strict subset
  $\vec{X}'$ of $\vec{X}$ such that $\vec{X}' = \vec{x}'$ satisfies
  EX1(a) and EX1(b)
  , where $\vec{x}'$ is the restriction of
$\vec{x}$ to the variables in $\vec{X}$.   
\item[{\rm EX3.}]
For some $(M,\vec{u}) \in \K$, we have that $(M,\vec{u}) \sat \vec{X}
= \vec{x} \land \phi$.  
(The agent considers possible a setting 
where the explanation and explanandum both hold.) 
\end{description}

The explanation is \emph{nontrivial} if it satisfies in 
\begin{description}
\item[{\rm EX4.}]
$(M',\vec{u}') \sat \neg(\vec{X}=\vec{x})$ for some $(M',\vec{u}' \in
  \K$
  such that $(M',\vec{u}') \sat \phi$.
(The explanation is not already known given the observation of
  $\phi$.)
\eprf
\end{description}
\end{definition}

The key part of the definition is EX1(b).  Roughly speaking, it says
that the explanation $\vec{X} = \vec{x}$ is a \emph{sufficient cause}
for $\phi$: for all settings that the agent considers possible,
intervening to set $\vec{X}$ to $\vec{x}$ results in $\phi$.  (See
\cite[Chapter 2.6]{Hal48} for a formal definition of sufficient cause.)
EX2, EX3, and EX4 should be fairly clear.  That leaves EX1(a).
Roughly speaking, it says that the explanation causes the
explanandum.  But there is a tension between EX1(a) and EX1(b) here:
we may need to add conjuncts to the explanation to ensure that it
suffices to make $\phi$ true in all contexts (as required by EX1(b)).
But these extra conjuncts may not be necessary to get causality in all contexts.
We want at least one of the conjuncts of $\vec{X} = \vec{x}$ to be
part of a cause of $\phi$, but we are willing to allow the cause to
include extra conjuncts.  To understand why, it is perhaps best to
look at an example.

\begin{example}
Going back to the forest-fire example,
consider the following four contexts: in $u_0 = (0,0)$, there is no lightning
and no arsonist; in $u_1 = (1,0)$, there is only lightning; in
$u_2 = (0,1)$, the arsonist drops a match but there is no lightning; and in
$u_3 = (1,1)$, there is lightning and the arsonist drops a match.
Let $M^d$ be the disjunctive model (where either lightning or a match
suffices to start the forest fire), and let $M^c$ be the conjunctive
model (where we need both).  Let 
$\K_1 = \{(M^d,u_0),(M^d,u_1),(M^d,u_2),(M^d,u_3)\}$.  Both $L=1$
and $\ML = 1$ are explanations of $\FF=1$ relative to $\K_1$.
Clearly EX1(b), EX2, EX3, and EX4 hold.  For EX1(a), recall that in
the setting $(M^d,u_3)$, the actual cause is $L=1 \land \ML = 1$.
Thus, EX1(a) is satisfied by $L=1$ by taking $\vec{Y} = \vec{y}$ to be
$\ML=1$, and is satisfied by $\ML=1$ by taking $\vec{Y} = y$ to br
$L=1$.

Now consider $\K_2 = \{(M^c,u_0),(M^c,u_1),(M^c,u_2),(M^c,u_3)\}$.
The only explanation of fire  relative to $\K_2$ is $L=1 \land
\ML=1$; due to the sufficiency requirement EX1(b), we need both conjuncts.  

To take just one more example,
if  $\K_3 = \{(M^c,u_1), (M^c,u_3)\}$, then
$\ML=1$ is an  explanation of the forest fire.
Since $L=1$ is already 
known, $\ML=1$ is all the additional information that the agent needs
to explain the fire. 
This is a trivial explanation: 
since both $\ML=1$ and $L=1$ are required for there to be a fire, the
agent knows $\ML=1$ when he see the fire.  Note that 
$\ML=1 \land L=1$ 
is not an explanation; it violates the minimality condition EX2.
$L=1$ is not an explanation either, since $(M^c,u_1)
\sat \neg [L\gets 1](\FF=1)$, so sufficient causality does not hold.
\exam

\subsection{Partial explanations and better explanations}
Not all explanations are considered equally good.  Moreover, it may be
hard to find an explanation that satisfies EX1 for all settings $(M,
\vec{u}) \in \K$; we may be satisfied with a formula that satisfies
these conditions for almost settings.  In \cite{Hal48},
various dimensions along which one explanation might be better than
another are discussed. We
focus on two of them here.  Here the probability $\Pr$ in the
epistemic state $(\K,\Pr)$ comes into
play. Given a causal formula $\phi$, let $\intension{\phi}_\K =
\{(M,\vec{u}) \in \K: (M,\vec{u}) \sat \phi\}$.  
If both $\vec{X} = \vec{x}$ and $\vec{X}' = \vec{x}'$ are
explanations of $\phi$ relative to $\K$, $\vec{X} = \vec{x}$ is
preferred relative to epistemic state $(\K,\Pr)$ if 
$\Pr(\intension{\vec{X} = \vec{x}}_\K \mid \intension{\phi}_\K) >
\Pr(\intension{\vec{X}' = \vec{x}'}_\K \mid \intension{\phi}_\K)$,
that is, if its prior probability is higher.
For example, if $\Pr(\{(M^d,u_1), (M^d,u_3)\}) > \Pr(\{(M^d,u_2),
    (M^d,u_3)\})$, 
    then $L=1$ would be viewed as a better explanation than $\ML=1$
    along this dimension; 
    it is more likely.
    
Another consideration takes as its point of departure the fact that
the conditions EX1(a) and (b) are rather stringent.  We might consider
$\vec{X}=\vec{x}$ quite a good explanation of $\phi$ relative to $\K$
if, with high probability, these conditions hold. More precisely, for
EX1(a), we
are interested in the probability of the set of settings $(M,\vec{u})$
in $\K$ for which there 
there exists a conjunct $X=x$ of \mbox{$\vec{X} = \vec{x}$} and a
(possibly empty)   conjunction $\vec{Y} = 
  \vec{y}$ such that \mbox{$X = x \land \vec{Y} = \vec{y}$} is
  a cause of $\phi$ in $(M,\vec{u})$, conditional on $\vec{X} =
  \vec{x} \land \phi$; for EX1(b), we are interested in the
  probability of the set of settings $(M,\vec{u}) \in \K$ for which
  $(M,\vec{u}) \sat [\vec{X} \gets \vec{x}]\phi$.  The higher these
  probabilities, the better the explanation.  Moreover, we can talk
  about a \emph{partial explanation}, one which satisfies EX1 with
  a probability less than 1.

\section{Dealing with mathematical explanations}\label{sec:mathexplanation}

We can now return to the questions that motivated this paper.  If we are
looking for a (causal) explanation of ``Why is 4373 the sum of two
squares?'', we need to start with a causal model.  If the explanation
is going to be ``it is a prime congruent to 1 mod 4'', the model needs
to include variables $P_{4373}$, $1M4_{4373}$, $S2S_{4373}$, where the first has
value 1 if 4373 is an odd prime and 0 otherwise; the
second has value 1 if 4373 is is congruent to 1 mod 4, and 0
otherwise; and the third has value 1 if 4373 is the sum of two squares
and 0 otherwise.

Now it is a fact of mathematics that 4373 is prime,
congruent to 1 mod 4, and the sum of two squares, but since the agent
does not necessarily know that (under some reasonable interpretation
of the word ``know''), we want to the agent to be able to consider
models where at least $P_{4373}$ and $S2S_{4373}$ take on value 0.
(We are implicitly assuming that since computing whether a number is
congruent to 1 mod 4 is so simple that the agent does in fact know that
$1M4_{4373} = 1$; 
nothing in the following discussion would change if we assumed that
the agent did not know this.)
We assume that the value of $P_{4373}$ and $IM4_{4373}$ is determined
by an exogenous variable $U$ that takes on four possible
values of the form $(i,j)$, where $i, j \in \{0,1\}$, as in the
forest-fire example; the value of $i$ determines the value of
$P_{4373}$ and the value of $j$ determines the value of $1M4_{4373}$.

If we further assume that the agent
knows Fermat's two-squares theorem,
then in all models that the agent considers
possible, if $P_{4373} = 1$ and $M4_{4373} = 1$, then $S2S_{4373} = 1$, and if
$P_{4373} = 1$ and $1M4_{4373} = 0$, then $S2S_{4373} = 0$.
The question is what should happen if $P_{4373} = 0$.  While it is not
hard to show that no number congruent to 3 mod 4 is the sum of two
squares, there are non-primes congruent to 0, 1, and 2 mod 4,
respectively, that 
are the sum of two squares (like 8, 25, and 18, respectively), and
non-primes congruent  to 0, 1, and 2 mod 4 that are not the sum of two
squares (like 12, 33, and 6, respectively).  
%1 mod 4 and are the sum of two
%squares (like 45) and nonprimes congruent to 1 mod 4 that are not the
%sum of two squares (like 33).  The only even prime (2) is clearly the
%sum of squares.  Finally, there are non-primes congruent to 3 mod 4
%that are not the sum of squares (like 15)  and non-primes congruent to
%3 mode 4 that are the sum of squares

It seems reasonable to assume that the
agent is uncertain about the effect of setting $P_{4373}$ to 0 on
$S2S$.  (Recall that although this would result in an ``impossible''
possible world, the agent still considers it possible.)
Thus, we consider two causal models that differ only in what 
happens if  $P_{4373}=0$: in one of them, call it $M_1$, it
results in $S2S_{4373} = 0$; in the other, call it $M_2$, results in
$S2S_{4373} = 1$.  (Otherwise, the models are identical.)
Let $u_0$, $u_1$, $u_2$, and $u_3$ be the
contexts where $U$ takes on values $(0,0)$, $(0,1)$, $(1,0)$, and
$(1,1)$, respectively.  Since we are assuming that the agent knows
that $1M4_{4373} = 1$, we can take $\K = \{(M_1,u_1), (M_1,u_3),
(M_2,u_1), (M_2,u_3)\}$.

It is now straightforward to show that, relative to $\K$, the fact that
$P_{4373} = 1 \land 1M4_{4373}=1$ (i.e., the fact the 4373 is an odd prime and
congruent to 1 mod 4) is an explanation of
the fact that 4373 is the sum of two squares.  Let us go  through the
conditions in Definition~\ref{def:explanation1}.  For EX1(a), there are
only two settings in $\K$ that satisfy $P_{4373} = 1 \land
1M4_{4373}=1$, namely $(M_1,u_3)$ and $(M_2,u_3)$.  In both of these
settings, $1M4_{4373}=1$ is a cause of $S2S_{4373} = 1$.  It is in
fact a but-for cause:  if we set  $1M4_{4373}=0$, then $S2S_{4373} =
0$ (since $P_{4373} = 1$ continues to hold).   EX1(b) is immediate,
since if 4373 is an odd prime that is congruent to 1 mod 4, then it is
guaranteed to be the sum of two squares (the equations in $M_1$ and
$M_2$ enforce this).  For EX2, note that $P_{4373} = 1$ does not
satisfy EX1(a) and $1M4_{4373}=1$ does not satisfy EX1(b).   EX3
clearly holds, since $P_{4373} = 1 \land
1M4_{4373}=1 \land S2S_{4373}$ holds in both $(M_1,u_3)$ and
$(M_2,u_3)$.  Finally, EX4 also holds; $\neg(P_{4373} = 1 \land
1M4_{4373}=1)$ holds in $(M_1,u_1)$ and $(M_2,u_1)$.

Although $1M4_{4373}=1$ is known (it is true in all settings
in $\K$), it is part of the explanation.  In general, an explanation
will not include facts that an agent already knows unless these facts
are necessary to show causality (i.e., EX1(a)).   Nothing would have
changed if $1M4_{4373}=1$ were not known (i.e., had we included
$(M_1,u_0)$, $(M_1,u_2)$, $(M_2,u_0)$, and $(M_2,u_2)$ in $\K$):
$P_{4373} = 1 \land 1M4_{4373}=1$ would still have been an explanation
of $S2S_{4373}$.

Now consider the other possible explanation of 4373 being the sum of
two squares, namely the demonstration that $4373 = 62^2 + 23^2$.  We
could capture this by adding another variable to the model,
$S2S_{4373,62,23}$, such that $S2S_{4373,62,23} = 1$ if $4373 = 62^2 +
23^2$ and 0 otherwise.  Again, we would allow the agent to consider it
possible that $S2S_{4373,62,23} = 0$.  We would expect that setting 
$S2S_{4373,62,23} = 1$ forces $S2S_{4373} = 1$, independent of the
values of the other variables (although, in light of the theorem,
we would not expect the agent to consider possible a model where
$S2S_{4373,62,23} = 1$, $P_{4373} = 1$, and $1M4_{4373} = 0$).  
But what if $S2S_{4373,62,23} = 0$?   We assume that in this
case, the agent is in the same situation he was in before, and
considers all the settings in $\K$ possible.  More precisely,
let $M_1'$ and $M_2'$ be like $M_1$ and $M_2$, respectively, except
that (1) the exogenous variable $U$ now has values of the form
$(i,j,k)$, for $i, j, k \in \{0,1\|$, and determines $P_{4373}$,
$1M4_{4373}$, and  
$S2S_{4373,62,23}$ in the obvious way; and (2) the equation for
$S2S_{4373}$ is such that if $S2S_{4373,62,23} = 1$, then
$S2S_{4373}=1$ in both 
models, while if $S2S_{4373,62,23} = 0$, then the value of $S2S_{4373}$ is
determined in $M_1'$ (resp., $M_2'$) in the same way that it is
determined in $M_1$ (resp., $M_2$).  

Let $u_0', \ldots, u_7'$ be the
contexts where $U'$ takes on values $(0,0,0)$, $(0,1,0)$, $(1,0,0)$, 
$(1,1,0)$, $(0,0,1)$, $(0,1,1)$, $(1,0,1)$,  and $(1,1,1)$, respectively;
let $\K' = \{(M',u'): M' \in \{M_1', M_2'\}, u'  \in
\{u_1',u_3',u_5',u_7'\}\}$. 
$S2S_{4373,62,23} = 1$ is not an explanation of $S2S_{4373}$ relative
to $\K'$ because it fails EX1(a).  For example, $S2S_{4373,62,23} =
1$ is not cause of $S2S_{4373} = 1$ in $(M_2',u_7')$.

The analysis of the second example is similar.
The fact that $x-2$
is a factor of $f(x)$ is an explanation of the fact that $f(2)=0$ (in
the model where there is a variable $F_{x-2}$ that is 1 if $x-2$ ia a
factor of $f(x)$ and 0 otherwise and a variable $f2E0$ that is 1 if
$f(2) = 0$ and 0 otherwise).  It seems reasonable to expect the
agent to understand that $x-2$ is factor of $f(x)$ iff $f(2) = 0$,
and for the equation for $f2E0$ to reflect this, so the fact that
$x-2$ is a factor of $f$ is a cause of $f2E0 = 1$.  The situation for
(\ref{eq:factor}) is a little more subtle.  It is not immediately
transparent that the left-hand side of (\ref{eq:factor}) is the result
of plugging in 2 for $x$ in the polynomial $f$.  Thus, the agent might
consider it possible that (\ref{eq:factor}) not hold and yet $f(2) = 0$;
more precisely, the agent would consider possible a model 
where (\ref{eq:factor}) does not hold and $f(2) = 0$.  In this model,
(\ref{eq:factor}) is not a cause of $f(2) = 0$, so if the agent's set of
possible settings includes this model, (\ref{eq:factor}) would not be
an explanation of $f(2)= 0$.  We would argue that at least part of the
reason that people find the fact that $x-2$ is a factor of $f$ a
better explanation for $f(2) = 0$ than (\ref{eq:factor}) is because
the fact that $f(2) = 0$ is not immediately obvious from
(\ref{eq:factor}).  

Now consider the quality of the explanation.  Although, as we have argued, 
the fact that $4373 = 62^2 + 23^2$, that is, $S2S_{4373,62,23} = 1$,
is not an explanation of 4373
being the sum of two squares, it could be a good partial explanation.
It does satisfy EX1(b) in all settings in $\K$, but satisfies EX1(a) only in
$(M_1',u_5')$, so how good a partial explanation it is depends on the
probability of $(M_1',u_5')$.  By way of contrast, $P_{4373}=1 \land
1M4_{43u73}=1$ 
satisfies EX1(a) and EX1(b) in all settings, so no matter what $\Pr$
is, it is at least as good an explanation as $S2S_{4373,62,23} = 1$,
and a strictly better explanation, at least as far as this
consideration goes, if $\Pr(M_1',u_5') < 1$.

On the other hand, if we consider the prior probability,
the probability of
$P_{4373}=1 \land 1M4_{43u73}=1$ conditional on $S2S_{4373} = 1$ is
not necessarily higher than the probability of $S2S_{4373,62,23} = 1$
conditional on $S2S_{4373} = 1$; it depends on the relative
probability of $\{(M_1',u_3'), (M_1',u_7'), (M_2',u_3'),
(M_2',u_7')$\} and $\{(M_2',u_5'), M_2',u_7')\}$.  While we have put
no constraints on $\Pr$, there is at least a heuristic argument that
the agent should consider the former event more likely than the
latter: by the prime number theorem [some reference], there are
roughly $\ln(n)$ prime numbers less than $n$; we would expect roughly
half of them to be 1 mod 4.  On the other hand, there is at most one
number less than $n$ that is $62^2 + 23^2$.  If $\Pr$ respects this
reasoning, then again, $P_{4373}=1 \land 1M4_{4373}=1$ is a better
explanation of $S2S_{4373} = 1$ than $S2S_{4373,62,23} = 1$.

In the second example, considering EX1(a) and EX1(b),
arguing just as in the first example, the fact that $x-2$
is a factor of $f$ is clearly at least as good an explanation as
(\ref{eq:factor}), and a strictly better explanation if the agent
places positive probability on a setting where (\ref{eq:factor}) does not
hold and $f(2) = 0$.   But now considering
the prior probability, conditional on $f(2) = 0$, the fact that $x-2$ is a
factor of $f$ has probability 1 (since the agent is assumed to know
that $f(2) =0$ iff $x-2$ is a factor of $f$), so again it is at least
as good an explanation when considering prior probability as
(\ref{eq:factor}), and a strictly better explanation if the  
agent places positive probability on a setting where (\ref{eq:factor})
does not hold and $f(2) = 0$.

\paragraph{Acknowledgments:} I thank Jordan Ellenburg for asking the
question that motivated this paper, and for useful conversations on
the subject of mathematical explanations.

\bibliographystyle{chicago}
\bibliography{z,joe}
\end{document}